\documentclass{article}

\usepackage{color}

\usepackage{subcaption}
\usepackage{placeins}

\usepackage{times}
\usepackage{epsfig}
\usepackage{amsmath}
\usepackage{amssymb}
\usepackage{bm}
\usepackage{amsthm}
\usepackage{multirow}
\usepackage{caption}
\usepackage[super]{nth}

\usepackage{amssymb}%
\usepackage{mathrsfs}%
\usepackage{nicefrac}
\usepackage{xcolor}
\usepackage{pifont}
\usepackage{array}
\usepackage{wrapfig}
\usepackage{mathtools}
\usepackage{colortbl}
\usepackage{mdframed}
\usepackage{enumitem}
\usepackage{soul}

\usepackage{microtype}
\usepackage{graphicx}
\usepackage{booktabs} 

\usepackage{hyperref}

\usepackage[accepted]{icml2023}

\usepackage{amsmath}
\usepackage{amssymb}
\usepackage{mathtools}
\usepackage{amsthm}

\usepackage[capitalize,noabbrev]{cleveref}

\theoremstyle{plain}

\theoremstyle{definition}

\theoremstyle{remark}

\usepackage[textsize=tiny]{todonotes}

\DeclareMathOperator*{\argmin}{arg\,min}
\DeclareMathOperator{\sgn}{sign}

\DeclareMathOperator{\clip}{clip}

\hypersetup{
	colorlinks=true,
	urlcolor=magenta,
}

\icmltitlerunning{Better Diffusion Models Further Improve Adversarial Training}

\begin{document}

\twocolumn[
\icmltitle{Better Diffusion Models Further Improve Adversarial Training}



\icmlsetsymbol{equal}{*}

\begin{icmlauthorlist}
\icmlauthor{Zekai Wang}{equal,to1}
\icmlauthor{Tianyu Pang}{equal,to2}
\icmlauthor{Chao Du}{to2}
\icmlauthor{Min Lin}{to2}
\icmlauthor{Weiwei Liu}{to1}
\icmlauthor{Shuicheng Yan}{to2}
\end{icmlauthorlist}

\icmlaffiliation{to1}{School of Computer Science, National Engineering Research Center for Multimedia Software, Institute of Artificial Intelligence and Hubei Key Laboratory of Multimedia and Network Communication Engineering, Wuhan University.}
\icmlaffiliation{to2}{Sea AI Lab}

\icmlcorrespondingauthor{Tianyu Pang}{tianyupang@sea.com}
\icmlcorrespondingauthor{Weiwei Liu}{liuweiwei863@gmail.com}

\icmlkeywords{Machine Learning, ICML}

\vskip 0.3in
]



\printAffiliationsAndNotice{$^{*}$Equal contribution. Work done during Zekai Wang's internship at Sea AI Lab} 

\begin{abstract}
    It has been recognized that the data generated by the denoising diffusion probabilistic model (DDPM) improves adversarial training. After two years of rapid development in diffusion models, a question naturally arises: can better diffusion models further improve adversarial training? This paper gives an affirmative answer by employing the most recent diffusion model \cite{conf/nips/Karras22edm} which has higher efficiency ($\sim 20$ sampling steps) and image quality (lower FID score) compared with DDPM. Our adversarially trained models achieve state-of-the-art performance on RobustBench using only generated data (no external datasets). Under the $\ell_\infty$-norm threat model with $\epsilon=8/255$, our models achieve $70.69\%$ and $42.67\%$ robust accuracy on CIFAR-10 and CIFAR-100, respectively, i.e. improving upon previous state-of-the-art models by $+4.58\%$ and $+8.03\%$. Under the $\ell_2$-norm threat model with $\epsilon=128/255$, our models achieve $84.86\%$ on CIFAR-10 ($+4.44\%$). These results also beat previous works that use external data. We also provide compelling results on the SVHN and TinyImageNet datasets. Our code is at \href{https://github.com/wzekai99/DM-Improves-AT}{https://github.com/wzekai99/DM-Improves-AT}.
\end{abstract}

\begin{table}[t]
\centering
\setlength{\tabcolsep}{3pt}
\vspace{-0.cm}
\caption{A brief summary comparison of test accuracy (\%) between our models and existing Rank \#1 models, \emph{with} (\ding{51}) and \emph{without} (\ding{55}) external datasets, as listed in RobustBench~\citep{conf/nips/CroceASDFCM021robustbench}. All of these models use the WRN-70-16 architecture.}
\vspace{0.2cm}
\renewcommand*{\arraystretch}{1.1}
    \begin{tabular}{ccccc}
        \toprule
        Dataset & Method & External   & Clean & AA    \\ \midrule
        \multirow{3}{*}{\begin{tabular}{c}\textbf{CIFAR-10}\\
    ($\ell_{\infty}$, $\epsilon=8/255$)
                \end{tabular}} &  \multirow{2}{*}{Rank \#1} & \ding{55} & 88.74 & 66.11 \\
                &  & \ding{51} & 92.23 & 66.58 \\
                \cmidrule{2-5}
                & \textbf{Ours} & \ding{55} & \textbf{93.25} & \textbf{70.69} \\

        \midrule
        \multirow{3}{*}{\begin{tabular}{c}\textbf{CIFAR-10}\\
    ($\ell_{2}$, $\epsilon=128/255$)
                \end{tabular}} &  \multirow{2}{*}{Rank \#1} & \ding{55} & 92.41 & 80.42 \\
                &  & \ding{51} & \textbf{95.74} & 82.32 \\
                \cmidrule{2-5}
                & \textbf{Ours} & \ding{55} & 95.54 & \textbf{84.86} \\

    \midrule
        \multirow{3}{*}{\begin{tabular}{c}\textbf{CIFAR-100}\\
    ($\ell_{\infty}$, $\epsilon=8/255$)
                \end{tabular}} &  \multirow{2}{*}{Rank \#1} & \ding{55} & 63.56 & 34.64 \\
                &  & \ding{51} & 69.15 & 36.88 \\
                \cmidrule{2-5}
                & \textbf{Ours} & \ding{55} & \textbf{75.22} & \textbf{42.67} \\
        \bottomrule
    \end{tabular}
\vspace{-0.cm}
\label{tab:sota}
\end{table}%

\vspace{-0.3cm}
\section{Introduction}
Adversarial training (AT) was first developed by \citet{Goodfellow2014}, which has proven to be one of the most effective defenses against adversarial attacks~\citep{conf/iclr/MadryMSTV18pgd,conf/icml/ZhangYJXGJ19trades,conf/icml/RiceWK20overfit} and dominated the winner solutions in adversarial competitions~\citep{kurakin2018competation,brendel2020adversarial}. 
It is acknowledged that the availability of more data is critical to the performance of adversarially trained models~\citep{conf/nips/SchmidtSTTM18,stutz2019disentangling}. 
Thus, several pioneer efforts are made to incorporate external datasets into AT~\citep{hendrycks2019using,conf/nips/CarmonRSDL19unlabel,conf/nips/AlayracUHFSK19,najafi2019robustness,zhai2019adversarially,wang2019improving}, either in a fully-supervised or semi-supervised learning paradigm.


However, external datasets, even if unlabeled, are not always available. 
According to recent research, data generated by the denoising diffusion probabilistic model (DDPM)~\citep{ho2020denoising} can also significantly enhance both clean and robust accuracy of adversarially trained models, which is considered as a type of learning-based data augmentation~\citep{journals/corr/Rebuffi21fixing,conf/nips/GowalRWSCM21,rade2021helperbased,conf/iclr/SehwagMHD0CM22,conf/icml/PangLYZY22score}.
Because of its effectiveness, on CIFAR-10 and CIFAR-100~\citep{Krizhevsky2012}, the data generated by DDPM is used by all existing top-rank models (without external datasets) listed in RobustBench~\citep{conf/nips/CroceASDFCM021robustbench}.\footnote{https://robustbench.github.io}

\begin{figure*}[t]
\begin{center}
\vspace{0.3cm}
\includegraphics[width=2.05\columnwidth]{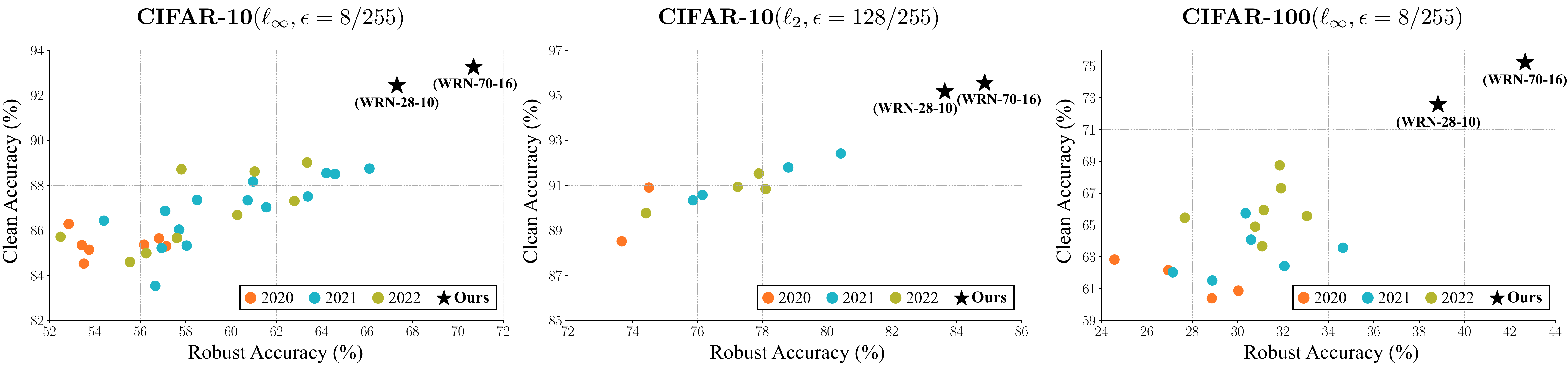}
\vspace{-0.3cm}
\caption{Robust accuracy (against AutoAttack) and clean accuracy of top-rank models (\emph{no external datasets}) in the leaderboard of RobustBench. The publication year of top-rank models is indicated by different colors. Our models use the WRN-28-10 and WRN-70-16 architectures in each setting, and detailed accuracy values are provided in Table~\ref{tab:sota_cifar10} and Table~\ref{tab:sota_cifar100}.}
\label{fig:1}
\end{center}
\vspace{-0.3cm}
\end{figure*}

After two years of rapid development in diffusion models, many improvements in sampling quality and efficiency have been made beyond the initial work of DDPM~\citep{conf/iclr/Song21scorebased}. 
In particular, the elucidating diffusion model (EDM)~\citep{conf/nips/Karras22edm} yields a new state-of-the-art (SOTA) FID score~\citep{conf/nips/HeuselRUNH17} of $1.97$ in an unconditional setting, compared to DDPM's FID score of $3.17$. 
While the images produced by DDPM and EDM are visually indistinguishable, we are curious whether better diffusion models (e.g., with lower FID scores) can benefit downstream applications even more (e.g., the task of AT).

It turns out that the reward for our curiosity is surprisingly good. 
We just replace the data generated by DDPM with the data generated by EDM and use almost the same training pipeline as described in \citet{journals/corr/Rebuffi21fixing}. 
As summarized in Table~\ref{tab:sota}, without the need for external datasets or additional training time per epoch, our adversarially trained WRN-70-16 models~\citep{conf/bmvc/ZagoruykoK16wrn} achieve new SOTA robust accuracy on CIFAR-10/CIFAR-100 under AutoAttack (AA)~\citep{conf/icml/Croce020autoattack}, associated with a large improvement in clean accuracy. Our models even surpass previous Rank \#1 models that rely on external data. The enhancements are significant enough---even our smaller model of WRN-28-10 architecture outperforms previous baselines, as shown in Figure~\ref{fig:1}. Our method can also substantially improve model performance on the SVHN and TinyImageNet datasets.

Moreover, we conduct extensive ablation studies to better reveal the mechanism by which diffusion models promote the AT process. Following the similar guidelines in \citet{gowal2020uncovering} and \citet{conf/iclr/PangYDSZ21bag}, we examine the effects of, e.g., quantity and quality of generated data, early stopping, and data augmentation. The results demonstrate that the data generated by EDM eliminates robust overfitting and reduces the generalization gap between clean and robust accuracy. During AT, we also conduct sensitivity analyses on a number of significant parameters. Our findings expand on the potential of learning-based data augmentation (i.e., using the data generated by diffusion models) and provide solid foundations for future research on promoting AT.

\section{Related Work}
\textbf{Diffusion models.} In recent years, denoising diffusion probabilistic modeling~\citep{sohl2015deep,ho2020denoising} and score-based Langevin dynamics~\citep{song2019generative,song2020improved} have shown promising results in image generation. \citet{conf/iclr/Song21scorebased} unify these two generative learning mechanisms using stochastic differential equations (SDE), and this unified model family is referred to as diffusion models. Later, there are emerging research routines that, to name a few, accelerate sampling inference~\citep{song2021denoising,lu2022dpm}, optimize model parametrization and sampling schedule~\citep{kingma2021variational,conf/nips/Karras22edm}, and adopt diffusion models in text-to-image generation~\citep{ramesh2022hierarchical,rombach2022high}.

\textbf{Adversarial training.} In addition to leveraging external datasets or generated data, several enhancements for AT have been made employing strategies inspired by other areas, including metric learning~\citep{mao2019metric,pang2019rethinking,pang2020boosting}, self-supervised learning~\citep{chen2020self,chen2020adversarial,naseer2020self,conf/icml/Wang022rvcl}, ensemble learning~\citep{tramer2017ensemble,pang2019improving}, fairness~\citep{conf/nips/MaW022, conf/aaai/Li23}, and generative modeling~\citep{jiang2018learning,wang2019direct,deng2020adversarial}. \citet{conf/nips/XuL22, conf/nips/LiXL22, journals/jmlr/Zou23} study adversarial robust learning from the theoretical perspective. Moreover, because of high computational cost of AT, various attempts have been made to accelerate the training phase by reusing calculation~\citep{shafahi2019adversarial,zhang2019you} or one-step training~\citep{wong2020fast,liu2020using,vivek2020single}. Some following studies address the side effects (e.g., catastrophic overfitting) induced by these fast AT approaches~\citep{andriushchenko2020understanding,li2020towards}.

\textbf{Adversarial purification.} Generative models have been used to purify adversarial examples~\citep{song2018pixeldefend} or strengthen certified defenses~\citep{carlini2022certified}. Diffusion models have recently gained popularity in adversarial purification~\citep{yoon2021adversarial,nie2022diffusion,wang2022guided,xiao2022densepure}, demonstrating promising robust accuracy against AutoAttack. The effectiveness of diffusion-based purification, on the other hand, is dependent on the randomness of the SDE solvers~\citep{ho2020denoising,bao2022analytic}, which causes at least tens of times inference computation and is unfriendly to downstream deployment. Furthermore, it has been demonstrated that stochastic pre-processing or test-time defenses have common limitations~\citep{gao2022limitations,croce2022evaluating}, which may be vulnerable to, e.g., transfer-based attacks~\citep{kang2021stable} or intermediate-state attacks~\citep{yang2022a}.

\textbf{Adversarial benchmarks.} Because of the large number of proposed defenses, it is critical to develop a comprehensive and up-to-date adversarial benchmark for ranking existing methods. \citet{dong2020benchmarking} perform large-scale experiments to generate robustness curves for evaluating typical defenses; \citet{tang2021robustart} provide comprehensive studies on how architecture design and training techniques affect robustness. Other benchmarks are available for specific scenarios, including adversarial patches~\citep{hingun2022reap,lian2022benchmarking,pintor2023imagenet}, language-related tasks~\citep{wang2021adversarial,li2021searching}, autonomous vehicles~\citep{xu2022safebench}, multiple threat models~\citep{hsiung2022carben}, and common corruptions~\citep{mu2019mnist,hendrycks2019benchmarking,sun2021certified}. In this paper, we use RobustBench~\citep{conf/nips/CroceASDFCM021robustbench}, which is a widely used benchmark in the community. RobustBench is built on AutoAttack, which has been proven to be reliable in evaluating deterministic defenses like adversarial training.

\begin{table*}[t]
\centering
\small
\vspace{-0.cm}
\caption{Test accuracy (\%) of clean images and under AutoAttack (AA) on CIFAR-10. We highlight our results in \textbf{bold} whenever the value represents an improvement relative to the strongest baseline using the same architecture, and we \underline{underline} them whenever the value achieves new SOTA result under the threat model. We did not apply CutMix following \citet{conf/icml/PangLYZY22score}. $^\dagger$With the same batch size, the training time per epoch of our method is equivalent to the w/o-generated-data baseline (see `training time' paragraph in \cref{sec:setup}).}
\vspace{0.1cm}
\renewcommand*{\arraystretch}{1.1}
\begin{tabular}{cclccccc}
\toprule
Dataset & Architecture & Method & Generated & Batch & Epoch$^\dagger$ & Clean & AA \\
\midrule
\multirow{19}{*}{\begin{tabular}{c}\textbf{CIFAR-10}\\
    ($\ell_{\infty}$, $\epsilon=8/255$)
                \end{tabular}} & WRN-34-20                  & \citet{conf/icml/RiceWK20overfit}                  & \ding{55}        & 128   & 200   & 85.34 & 53.42 \\
                    & WRN-34-10                  & \citet{conf/icml/ZhangXH0CSK20kill}                 & \ding{55}        & 128   & 120   & 84.52 & 53.51 \\
                    & WRN-34-20                  & \citet{conf/iclr/PangYDSZ21bag}                  & \ding{55}        & 128   & 110   & 86.43 & 54.39 \\
                    & WRN-34-10                  & \citet{conf/nips/WuX020}                    & \ding{55}        & 128   & 200   & 85.36 & 56.17 \\
                    & WRN-70-16                  & \citet{gowal2020uncovering}                 & \ding{55}        & 512   & 200   & 85.29 & 57.14 \\ 
                    & WRN-34-10                  & \citet{conf/iclr/SehwagMHD0CM22}                 & 10M        & 128   & 200   & 87.00 & 60.60 \\ \cmidrule{2-8} 
                    & \multirow{7}{*}{WRN-28-10} & \citet{journals/corr/Rebuffi21fixing}               & 1M        & 1024  & 800   & 87.33 & 60.73 \\
                    &                            & \citet{conf/icml/PangLYZY22score}                  & 1M        & 512   & 400   & 88.10  & 61.51 \\
                    &                            & \citet{conf/nips/GowalRWSCM21}              & 100M      & 1024  & 2000  & 87.50  & 63.38 \\ \cmidrule{3-8} 
                    &                            & \multirow{4}{*}{\textbf{Ours}} & 1M        & 512   & 400   & \textbf{91.12} & 63.35 \\
                    &                            &                       & 1M        & 1024  & 800   & \textbf{91.43} & \textbf{63.96} \\
                    &                            &                       & 50M       & 2048  & 1600  & \textbf{92.27} & \textbf{67.17} \\ 
                    &                            &                       & 20M       & 2048  & 2400  & \textbf{92.44} & \textbf{67.31} \\ \cmidrule{2-8} 
                    & \multirow{6}{*}{WRN-70-16} & \citet{conf/icml/PangLYZY22score}                  & 1M        & 512   & 400   & 88.57 & 63.74 \\
                    &                            & \citet{journals/corr/Rebuffi21fixing}               & 1M        & 1024  & 800   & 88.54 & 64.20  \\
                    &                            & \citet{conf/nips/GowalRWSCM21}              & 100M      & 1024  & 2000  & 88.74 & 66.11  \\ \cmidrule{3-8} 
                    &                            & \multirow{3}{*}{\textbf{Ours}} & 1M        & 512   & 400   & \textbf{91.98} & 65.54 \\
                    &                            &                       & 5M        & 512   & 800   & \textbf{92.58} & \textbf{67.92} \\
                    &                            &                       & 50M       & 1024  & 2000  & \underline{\textbf{93.25}}     & \underline{\textbf{70.69}}     \\ \midrule
\multirow{10}{*}{\begin{tabular}{c}\textbf{CIFAR-10}\\
    ($\ell_{2}$, $\epsilon=128/255$)
                \end{tabular}}  & WRN-34-10                  & \citet{conf/nips/WuX020}                    & \ding{55}        & 128   & 200   & 88.51 & 73.66 \\
                    & WRN-70-16                  & \citet{gowal2020uncovering}              & \ding{55}        & 512   & 200   & 90.9  & 74.50  \\ 
                    & WRN-34-10                  & \citet{conf/iclr/SehwagMHD0CM22}              & 10M        & 128   & 200   & 90.80  & 77.80  \\  \cmidrule{2-8} 
                    & \multirow{4}{*}{WRN-28-10} & \citet{conf/icml/PangLYZY22score}                  & 1M        & 512   & 400   & 90.83 & 78.10  \\
                    &                            & \citet{journals/corr/Rebuffi21fixing}               & 1M        & 1024  & 800   & 91.79 & 78.69 \\ \cmidrule{3-8} 
                    &                            & \multirow{2}{*}{\textbf{Ours}} & 1M        & 512   & 400   & \textbf{93.76} & \textbf{79.98} \\
                    &                            &                       & 50M       & 2048  & 1600  & \textbf{95.16}     & \textbf{83.63}     \\ \cmidrule{2-8} 
                    & \multirow{3}{*}{WRN-70-16} & \citet{journals/corr/Rebuffi21fixing}               & 1M        & 1024  & 800   & 92.41 & 80.42 \\ \cmidrule{3-8} 
                    &                            & \multirow{2}{*}{\textbf{Ours}} & 1M        & 512   & 400   & \textbf{94.47} & \textbf{81.16} \\
                    &                            &                       & 50M       & 1024  & 2000  & \underline{\textbf{95.54}}     & \underline{\textbf{84.86}}     \\ 
\bottomrule

\end{tabular}%
\label{tab:sota_cifar10}
\vspace{-0.2cm}
\end{table*}%

\section{Experiment Setup}\label{sec:setup}

We follow the basic setup and use the PyTorch implementation of \citet{journals/corr/Rebuffi21fixing}.\footnote{https://github.com/imrahulr/adversarial\_robustness\_pytorch} More information about the experimental settings can be found in \cref{app:setup}.


\textbf{Model architectures.} As our backbone networks, we adopt WideResNet (WRN)~\citep{conf/bmvc/ZagoruykoK16wrn} with the Swish/SiLU activation function~\citep{journals/corr/HendrycksG16}. We use WRN-28-10 and WRN-70-16, the two most common architectures on RobustBench~\citep{conf/nips/CroceASDFCM021robustbench}.

\textbf{Generated data.} To generate new images, we use the elucidating diffusion model (EDM)~\citep{conf/nips/Karras22edm} that achieves SOTA FID scores. We employ the \emph{class-conditional} EDM, whose training does not rely on external datasets (except for TinyImageNet, as specified in \cref{sec:main_results}). We follow the guidelines in \citet{conf/nips/CarmonRSDL19unlabel} to generate 1M CIFAR-10/CIFAR-100 images, which are selected from 5M generated images, with each image scored by a standardly pretrained WRN-28-10 model. We select the top $20\%$ scoring images for each class. When the amount of generated data exceeds 1M, or when generating data for SVHN/TinyImageNet, we adopt all generated images without selection. Note that unlike \citet{journals/corr/Rebuffi21fixing} and \citet{conf/nips/GowalRWSCM21} that use unconditional DDPM, the pseudo-labels of the generated images are directly determined by the class conditioning in our implementation.


\textbf{Training settings.} We use TRADES~\citep{conf/icml/ZhangYJXGJ19trades} as the framework of adversarial training (AT), with $\beta=5$ for CIFAR-10/CIFAR-100, $\beta=6$ for SVHN, and $\beta=8$ for TinyImageNet. We adopt weight averaging with decay rate $\tau=0.995$~\citep{conf/uai/IzmailovPGVW18wa}. We use the SGD optimizer with Nesterov momentum~\citep{nesterov1983method}, where the momentum factor and weight decay are set to $0.9$ and $5\times 10^{-4}$, respectively. We use the cyclic learning rate schedule with cosine annealing~\citep{smith2019super}, where the initial learning rate is set to $0.2$.

\textbf{Training time.} Regardless of the amount of generated data used (e.g., 1M, 20M, or 50M), the number of iterations per training epoch is controlled to be $\left \lceil{\frac{\textrm{amount of original data}}{\textrm{batch size}}}\right \rceil$ for all of our experiments utilizing generated data. This ensures a fair comparison with the w/o-generated-data baselines (e.g., those marked with \ding{55} in Table~\ref{tab:sota_cifar10}), as the training time remains constant when the number of training epochs and batch size are fixed. In particular, we sample images from the original and generated data in every training batch with a fixed `original-to-generated ratio'. Using CIFAR-10 (50K training images) and an original-to-generated ratio of 0.3, for example, each epoch involves training the model on 50K images: 15K from the original data and 35K from the generated data. For CIFAR-10/CIFAR-100 experiments, the original-to-generated ratio is $0.3$ for 1M generated data and $0.2$ when the required generated data exceeds 1M. \cref{tab:other_datasets} contains the original-to-generated ratios applied to SVHN and TinyImageNet, while \cref{app:split} contains additional ablation studies regarding the effects of different ratios.


\textbf{Evaluation metrics.} We evaluate model robustness against AutoAttack~\citep{conf/icml/Croce020autoattack}. Due to the high computation cost of AT, we cannot afford to report standard deviation for each experiment. For clarification, we train a WRN-28-10 model on CIFAR-10 with 1M generated data five times, using the batch size of $512$ and running for $400$ epochs. The clean accuracy is $91.12 \pm 0.15\%$, and the robust accuracy under the $(\ell_{\infty},\epsilon=8/255)$ threat model is $63.35 \pm 0.12\%$, indicating that our results have low variances.

\begin{table*}[t]
\centering
\small
\vspace{-0.cm}
\caption{Test accuracy (\%) of clean images and under AutoAttack (AA) on CIFAR-100. We highlight our results in \textbf{bold} whenever the value represents an improvement relative to the strongest baseline using the same architecture, and we \underline{underline} them whenever the value achieves new SOTA result under the threat model. We did not apply CutMix following \citet{conf/icml/PangLYZY22score}. $^\dagger$With the same batch size, the training time per epoch of our method is equivalent to the w/o-generated-data baseline (see `training time' paragraph in \cref{sec:setup}).}
\vspace{0.1cm}
\renewcommand*{\arraystretch}{1.1}
\begin{tabular}{cclccccc}
\toprule
Dataset & Architecture & Method & Generated & Batch & Epoch$^\dagger$ & Clean & AA \\
\midrule
\multirow{11}{*}{\begin{tabular}{c}\textbf{CIFAR-100}\\
    ($\ell_{\infty}$, $\epsilon=8/255$)
                \end{tabular}} & WRN-34-10       & \citet{conf/nips/WuX020}                    & \ding{55}        & 128   & 200   & 60.38 & 28.86 \\
                    & WRN-70-16                  & \citet{gowal2020uncovering}                 & \ding{55}        & 512   & 200   & 60.86 & 30.03 \\ 
                    & WRN-34-10                  & \citet{conf/iclr/SehwagMHD0CM22}                 & 1M        & 128   & 200   & 65.90 & 31.20 \\ \cmidrule{2-8} 
                    & \multirow{4}{*}{WRN-28-10} & \citet{conf/icml/PangLYZY22score}                  & 1M        & 512   & 400   & 62.08 & 31.40  \\
                    &                            & \citet{journals/corr/Rebuffi21fixing}               & 1M        & 1024  & 800   & 62.41 & 32.06 \\ \cmidrule{3-8} 
                    &                            & \multirow{2}{*}{\textbf{Ours}} & 1M        & 512   & 400   & \textbf{68.06} & \textbf{35.65} \\
                    &                            &                       & 50M       & 2048  & 1600  & \textbf{72.58}     & \textbf{38.83}    \\ \cmidrule{2-8} 
                    & \multirow{4}{*}{WRN-70-16} & \citet{conf/icml/PangLYZY22score}                  & 1M        & 512   & 400   & 63.99 & 33.65 \\
                    &                            & \citet{journals/corr/Rebuffi21fixing}               & 1M        & 1024  & 800   & 63.56 & 34.64 \\ \cmidrule{3-8} 
                    &                            & \multirow{2}{*}{\textbf{Ours}} & 1M        & 512   & 400   & \textbf{70.21}     & \textbf{38.69}     \\
                    &                            &                       & 50M       & 1024  & 2000  & \underline{\textbf{75.22}}     & \underline{\textbf{42.67}}     \\ 
\bottomrule
\end{tabular}%
\label{tab:sota_cifar100}
\vspace{-0.35cm}
\end{table*}%

\begin{table*}[t]
\centering
\small
\vspace{-0.cm}
\caption{Test accuracy (\%) of clean images and under AutoAttack (AA) on SVHN and TinyImageNet. We highlight the results following the notations in \cref{tab:sota_cifar100}. Here `Ratio' indicates the original-to-generated ratio. All the results adopt the WRN-28-10 model architecture. We did not apply CutMix following \citet{conf/icml/PangLYZY22score}. $^\dagger$Note that \citet{conf/nips/GowalRWSCM21} utilize the class-conditional DDPM model on ImageNet \cite{conf/nips/DhariwalN21beatgan} and directly generate images using the labels of TinyImageNet as the class condition.}
\vspace{0.1cm}
\renewcommand*{\arraystretch}{1.1}
\begin{tabular}{clcccccc}
\toprule
Dataset & Method & Generated & Ratio & Batch & Epoch & Clean & AA \\
\midrule
\multirow{4}{*}{\begin{tabular}{c}\textbf{SVHN}\\
    ($\ell_{\infty}$, $\epsilon=8/255$)
                \end{tabular}}                  & \citet{conf/nips/GowalRWSCM21} & \ding{55}   & \ding{55}  & 512   & 400   & 92.87	 & 56.83 \\
                    & \citet{conf/nips/GowalRWSCM21}                 & 1M    & 0.4    & 1024   & 800   & 94.15 & 60.90 \\ 
                    & \citet{journals/corr/Rebuffi21fixing}                 & 1M    & 0.4    & 1024   & 800   & 94.39 & 61.09 \\ 
                    & \multirow{2}{*}{\textbf{Ours}} & 1M    & 0.2    & 1024   & 800   & \textbf{95.19} & \textbf{61.85} \\
                    &                       & 50M   & 0.2    & 2048  & 1600  & \underline{\textbf{95.56}} & \underline{\textbf{64.01}} \\ 
                    \midrule

\multirow{4}{*}{\begin{tabular}{c}\textbf{TinyImageNet}\\
    ($\ell_{\infty}$, $\epsilon=8/255$)
                \end{tabular}}                  & \citet{conf/nips/GowalRWSCM21}                  & \ding{55}   & \ding{55}   & 512   & 400   & 51.56		 & 21.56 \\
                    & \textbf{Ours}                 & 1M   & 0.4     & 512   & 400   & \textbf{53.62} & \textbf{23.40} \\ \cmidrule{2-8} 
                    & \citet{conf/nips/GowalRWSCM21}$^\dagger$  & 1M    & 0.3    & 1024   & 800   & 60.95 & 26.66 \\
                    &  \textbf{Ours} (ImageNet EDM)                     & 1M    & 0.2   & 512  & 400  & \underline{\textbf{65.19}} & \underline{\textbf{31.30}} \\ 
\bottomrule

\end{tabular}%
\label{tab:other_datasets}
\vspace{-0.2cm}
\end{table*}%

\section{Comparison with State-of-the-Art}\label{sec:main_results}

We compare our adversarially trained models with top-rank models in RobustBench that do not use external datasets. \cref{tab:sota_cifar10} shows the results under the ($\ell_{\infty}$, $\epsilon=8/255$) and ($\ell_{2}$, $\epsilon=128/255$) threat models on CIFAR-10; \cref{tab:sota_cifar100,tab:other_datasets} presents the results under the ($\ell_{\infty}$, $\epsilon=8/255$) threat model on CIFAR-100, SVHN, and TinyImageNet. In summary, previous top-rank models use images generated by DDPM, whereas our models use EDM and significantly improves both clean and robust accuracy. Our best models beat all RobustBench entries (including those that use external datasets) under these threat models.

\textbf{Remark for \cref{tab:sota_cifar10}.} Under the ($\ell_{\infty}$, $\epsilon=8/255$) threat model on CIFAR-10, even when using 1M generated images, small batch size of $512$ and short training of $400$ epochs, our WRN-28-10 model achieves the robust accuracy comparable with \citet{conf/nips/GowalRWSCM21} that use 100M generated images, while our clean accuracy improves significantly ($+3.62\%$). After applying a larger batch size of $2048$ and longer training of $2400$ epochs, our WRN-28-10 model surpasses the previous best result obtained by 100M generated data with a large margin (clean accuracy $+4.94\%$, robust accuracy $+3.93\%$), and even beats previous SOTA of WRN-70-16 model. When using 50M generated images and training for $2000$ epochs on WRN-70-16, our model reaches $93.25\%$ clean accuracy and $70.69\%$ robust accuracy, obtaining improvements of $+4.51\%$ and $+4.58\%$ over the SOTA result, respectively. \ul{This is the first adversarially trained model to achieve clean accuracy over $90\%$ and robust accuracy over $70\%$ without external datasets.} Under the ($\ell_{2}$, $\epsilon=128/255$) threat model on CIFAR-10, our best WRN-28-10 model achieves $95.16\%$ ($+3.37\%$) clean accuracy and $83.63\%$ ($+4.94\%$) robust accuracy; our best WRN-70-16 model achieves $95.54\%$ ($+3.13\%$) clean accuracy and $84.86\%$ ($+4.44\%$) robust accuracy, which improve noticeably upon previous SOTA models.

\textbf{Remark for \cref{tab:sota_cifar100}.} Using the images generated by EDM under the ($\ell_\infty$, $\epsilon=8/255$) threat model on CIFAR-100 results in surprisingly good performance. Specifically, our best WRN-28-10 model achieves $72.58\%$ ($+10.17\%$) clean accuracy and $38.83\%$ ($+6.77\%$) robust accuracy; our best WRN-70-16 model achieves $75.22\%$ ($+11.66\%$) clean accuracy and $42.67\%$ ($+8.03\%$) robust accuracy.

\textbf{Remark for \cref{tab:other_datasets}.} We evaluate performance under the ($\ell_{\infty}$, $\epsilon=8/255$) threat model on SVHN and TinyImageNet datasets. As seen, our approach significantly outperforms the baselines. Our best model achieves $64.01\%$ ($+2.92\%$) robust accuracy on SVHN using 50M generated data.

We train the EDM model exclusively on TinyImageNet's training set to produce 1M data. Our WRN-28-10 model obtains improvements of $+2.06\%$ and $+1.84\%$ over clean and robust accuracy, respectively.  Notably, \citet{conf/nips/GowalRWSCM21} utilize the class-conditional DDPM pre-trained on ImageNet \cite{conf/nips/DhariwalN21beatgan} for data generation on TinyImagenet, since TinyImagenet dataset is a subset of ImageNet. To ensure a fair comparison, we use the checkpoint pre-trained on ImageNet provided by EDM, and class-conditional generate the images with the specific classes of TinyImageNet. The improvements are remarkable: our model achieves $65.19\%$ ($+4.24\%$) clean accuracy and $31.30\%$ ($+4.64\%$) robust accuracy with a smaller batch size and epoch than \citet{conf/nips/GowalRWSCM21}. These results demonstrate the effectiveness of generated data in enhancing model robustness across multiple datasets.

\begin{table*}[t]
\small
\vspace{-0.1cm}
    \caption{Test accuracy (\%) when training for different \textbf{number of epochs}, under the ($\ell_{\infty}$, $\epsilon=8/255$) threat model on CIFAR-10. WRN-28-10 models are trained with the batch size of $2048$ and original-to-generated ratio $0.2$. The model achieves the highest PGD-40 accuracy on the validation set at the `Best epoch'. `Early' and `Last' mean the test performance at the best and last epoch, respectively. `Diff' denotes the accuracy gap between the `Early' and `Last'.}
    \vspace{0.1cm}
    \centering
    \renewcommand*{\arraystretch}{1.1}
    \begin{tabular}{cccccccccccccc}
    \toprule
    \multirow{2}*{Generated} & \multirow{2}*{Epoch} & \multirow{2}*{Best epoch} & \multicolumn{3}{c}{Clean} & & \multicolumn{3}{c}{PGD-40}  & & \multicolumn{3}{c}{AA}\\
    \cmidrule{4-6}
    \cmidrule{8-10}
    \cmidrule{12-14}
    & & & Early & Last & Diff & & Early & Last & Diff & & Early & Last & Diff \\
    \midrule
    \multirow{2}{*}{\ding{55}}  & 400   & 86         & 84.41 & 82.18 & $-$2.23 &  & 55.23  & 46.21  & $-$9.02    &  & 54.57 & 44.89 & $-$9.68  \\
    & 800   & 88         & 83.60  & 82.15 & $-$1.45 &  & 53.86 & 45.75 & $-$8.11 &  & 53.13 & 44.58 & $-$8.55  \\ \midrule
\multirow{6}{*}{20M} & 400   & 370        & 91.27 & 91.45 & $+$0.18  &  & 64.65 & 64.80  & $+$0.15  &  & 63.69 & 63.84 & $+$0.15   \\
    & 800   & 755        & 92.08 & 92.14 & $+$0.06  &  & 66.61 & 66.72 & $+$0.11  &  & 65.66 & 65.63 & $+$0.03      \\
    & 1200  & 1154       & 92.43 & 92.32 & $-$0.11 &  & 67.45 & 67.64 & $+$0.19  &  & 66.31 & 66.60  & $+$0.29   \\
    & 1600  & 1593       & 92.51 & \textbf{92.61} & $+$0.10  &  & 68.05 & 67.98 & $-$0.07 &  & 67.14 & 67.10  & $-$0.04  \\
    & 2000  & 1978       & 92.41 & 92.55 & $+$0.14  &  & 68.32 & 68.30 & $-$0.02     &  & 67.22 & 67.17 & $-$0.05  \\
    & 2400  & 2358       & \textbf{92.58} & 92.54 & $-$0.04 &  & \textbf{68.43} & \textbf{68.39} & $-$0.04 &  & \textbf{67.31} & \textbf{67.30}     & $-$0.01 \\
    \bottomrule
    \end{tabular}
    \label{tab:epoch}
    \vspace{-0.1cm}
\end{table*}

\section{How Generated Data Influence Robustness}\label{sec:gen_data}
\citet{conf/icml/RiceWK20overfit} first observe the phenomenon of robust overfitting in AT: the test robust loss turns into increasing after a specific training epoch, e.g., shortly after the learning rate decay. The cause of robust overfitting is still debated~\citep{conf/icml/PangLYZY22score}, but one widely held belief is that the dataset is not large enough to achieve robust generalization~\citep{conf/nips/SchmidtSTTM18}. When the training set is dramatically expanded, using a large amount of external data~\citep{conf/nips/CarmonRSDL19unlabel, conf/nips/AlayracUHFSK19} or synthetic data~\citep{conf/nips/GowalRWSCM21}, significant improvements in both clean and robust accuracy are observed. In this section, we comprehensively study how the training details affect robust overfitting and model performance \emph{when generated images are applied to AT}. Unless otherwise specified, the experiments are carried out on CIFAR-10 dataset and WRN-28-10 models that have been trained for $400$ epochs, with a batch size of $512$ and an original-to-generated ratio of $0.3$.

\subsection{Early Stopping and Number of Epochs}
In the standard setting, a line of research \citep{conf/iclr/ZhangBHRV17, journals/corr/Belkin19} has found that the deep learning model does not exhibit overfitting in practice, i.e., the testing loss decreases alongside the training loss and it is a common practice to train for as long as possible. For AT, however, \citet{conf/icml/RiceWK20overfit} reveal the phenomenon of robust overfitting: robust accuracy degrades rapidly on the test set while it continues to increase on the training set. A larger training epoch does not guarantee improved performance in the absence of generated data. Thus, early stopping becomes a default option in the AT process, which tracks the robust accuracy on a hold-out validation set and selects the checkpoint with the best validation robustness.

We conduct experiments on CIFAR-10 with 20M images generated by EDM to investigate how the training epoch and early stopping affect robust performance \emph{when sufficient images are used}. The results are displayed in \cref{tab:epoch}. We also provide results with no generated data for better comparison, and we can conclude that:

\begin{itemize}
    \item 
    Early stopping is effective when no generated data is used, as previously observed~\citep{conf/icml/RiceWK20overfit}. Early stopping is triggered during the training process's initial phase ($86$-th epoch/$400$ epochs; $88$-th epoch/$400$ epochs). On the test set, both clean and robust accuracy degrade, and longer training leads to poor performance.
    
    \item 
    Early stopping is less important when using generated data. The best-performing model appears at the end of the training, implying that stopping early will not result in a significant improvement. `Diff' becomes minor, indicating that adequate training data effectively mitigates robust overfitting. 
    
    \item 
    The model performs better with a longer training process when 20M generated images are used. Surprisingly, a short training epoch results in robust underfitting (`Diff' is positive). When the model is trained on enough data, the results suggest that training as long as possible benefits the robust performance.
    \vspace{-0.1cm}
\end{itemize}

We regard early stopping as a default trick consistent with previous works~\citep{conf/iclr/PangYDSZ21bag} because of its effectiveness on the original dataset and comparable performance on big data. Refer to \cref{app:setup} for implementation details. 


\begin{table*}[t]
    \small
        \caption{Test accuracy (\%) when trained with different \textbf{amounts of generated data}, under the ($\ell_{\infty}$, $\epsilon=8/255$) threat model on CIFAR-10. The model achieves the highest PGD-40 accuracy on the validation set at the `Best epoch'. `Best' is the highest accuracy ever achieved during training; `Last' is the test performance at the last epoch. `Diff' denotes the gap between `Best' and `Last'. Since running AA is time-consuming, we regard AA accuracy at `Best epoch' as the `Best'.}
        \centering
        \vspace{0.1cm}
        \renewcommand*{\arraystretch}{1.1}
        \begin{tabular}{ccccccccccccc}
        \toprule
        \multirow{2}*{Generated}  & \multirow{2}*{Best epoch}  & \multicolumn{3}{c}{Clean} & & \multicolumn{3}{c}{PGD-40}  & & \multicolumn{3}{c}{AA}\\
        \cmidrule{3-5}
        \cmidrule{7-9}
        \cmidrule{11-13}
        & & Best & Last & Diff & & Best & Last & Diff & & Best & Last & Diff \\
        \midrule
        \ding{55}         & 91         & 84.55 & 82.59 & $-$1.96 &  & 55.66 & 46.47 & $-$9.19 &  & 54.37 & 45.29 & $-$9.08 \\
        50K       & 171        & 86.15 & 85.47 & $-$0.68 &  & 56.96 & 50.02 & $-$6.94 &  & 55.71 & 48.85 & $-$6.86 \\
        100K      & 274        & 88.20 & 87.47 & $-$0.73 &  & 59.85 & 54.95 & $-$4.90  &  & 58.85 & 53.42 & $-$5.43 \\
        200K      & 365        & 89.71 & 89.48 & $-$0.23 &  & 61.69 & 60.32 & $-$1.37 &  & 59.91 & 59.11 & $-$0.80  \\
        500K      & 395        & 90.76 & 90.58 & $-$0.18 &  & 63.85 & 63.69 & $-$0.16 &  & 62.76 & 62.77 & $+$0.01  \\
        1M        & 394        & 91.13 & 90.89 & $-$0.24 &  & 64.67 & 64.50 & $-$0.17 &  & 63.35 & 63.50 & $+$0.15  \\
        5M        & 395        & 91.15 & 90.93 & $-$0.22 &  & 64.88 & 64.88 & 0     &  & 64.05 & 64.05 & 0     \\
        10M       & 396        & \textbf{91.25} & \textbf{91.18} & $-$0.07 &  & 65.03 & 64.96 & $-$0.07 &  & 64.19 & 64.28 & $+$0.09  \\
        20M       & 399        & 91.17 & 91.07 & $-$0.10 &  & 65.21 & 65.13 & $-$0.08 &  & 64.27 & 64.16 & $-$0.11 \\
        50M       & 395        & 91.24 & 91.15 & $-$0.09 &  & \textbf{65.35} & \textbf{65.23} & $-$0.12 &  & \textbf{64.53} & \textbf{64.51} & $-$0.02 \\
        \bottomrule
        \end{tabular}
        \label{tab:amount}
        \vspace{-0.cm}
    \end{table*}
    
\begin{figure*}[t]
    \centering
    \vspace{-0.15cm}
    \begin{subfigure}[t]{0.247\linewidth}
    \includegraphics[width=\linewidth]{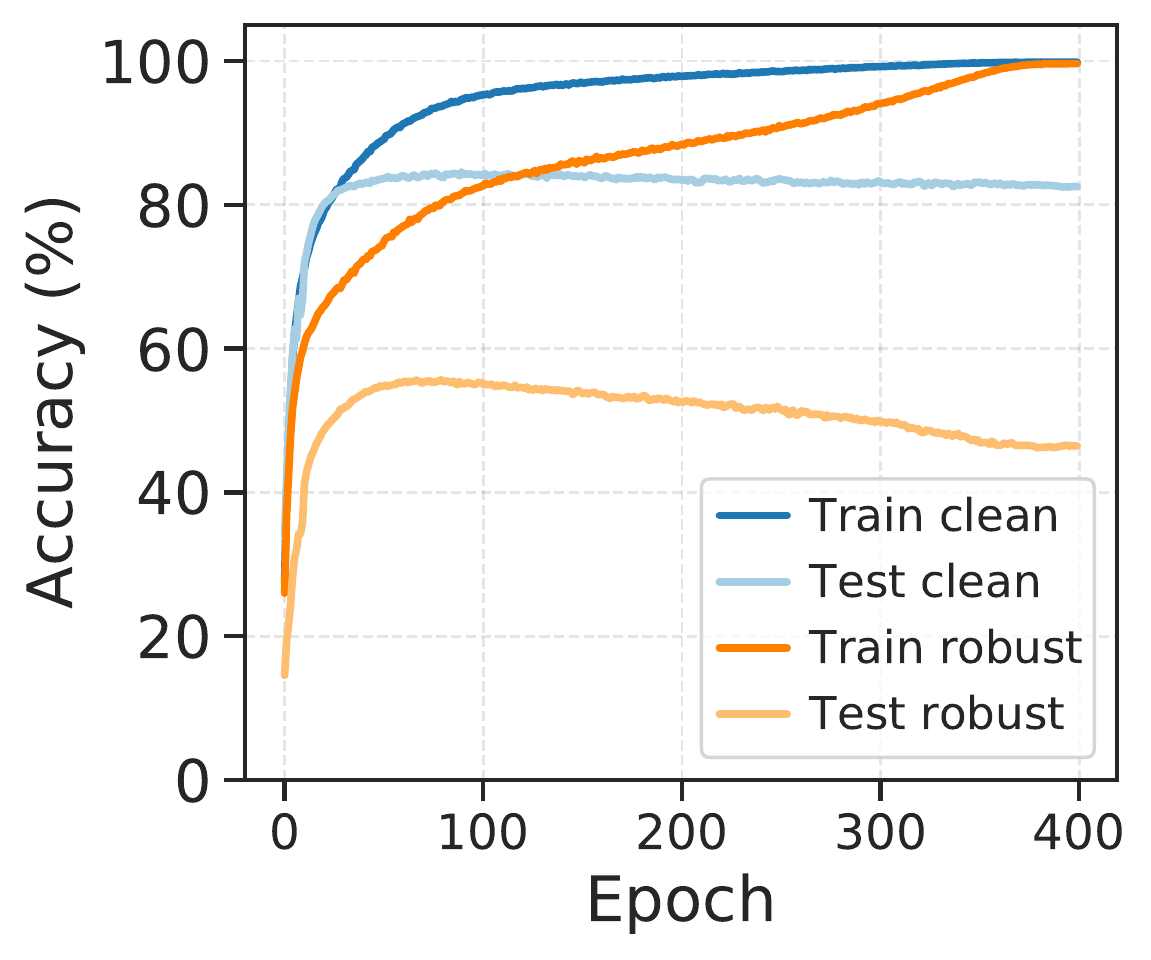}
    \vskip -0.08in
    \caption{no generated data}
    \label{fig:amount_no}
  \end{subfigure}	
  \begin{subfigure}[t]{0.247\linewidth}
        \includegraphics[width=\linewidth]{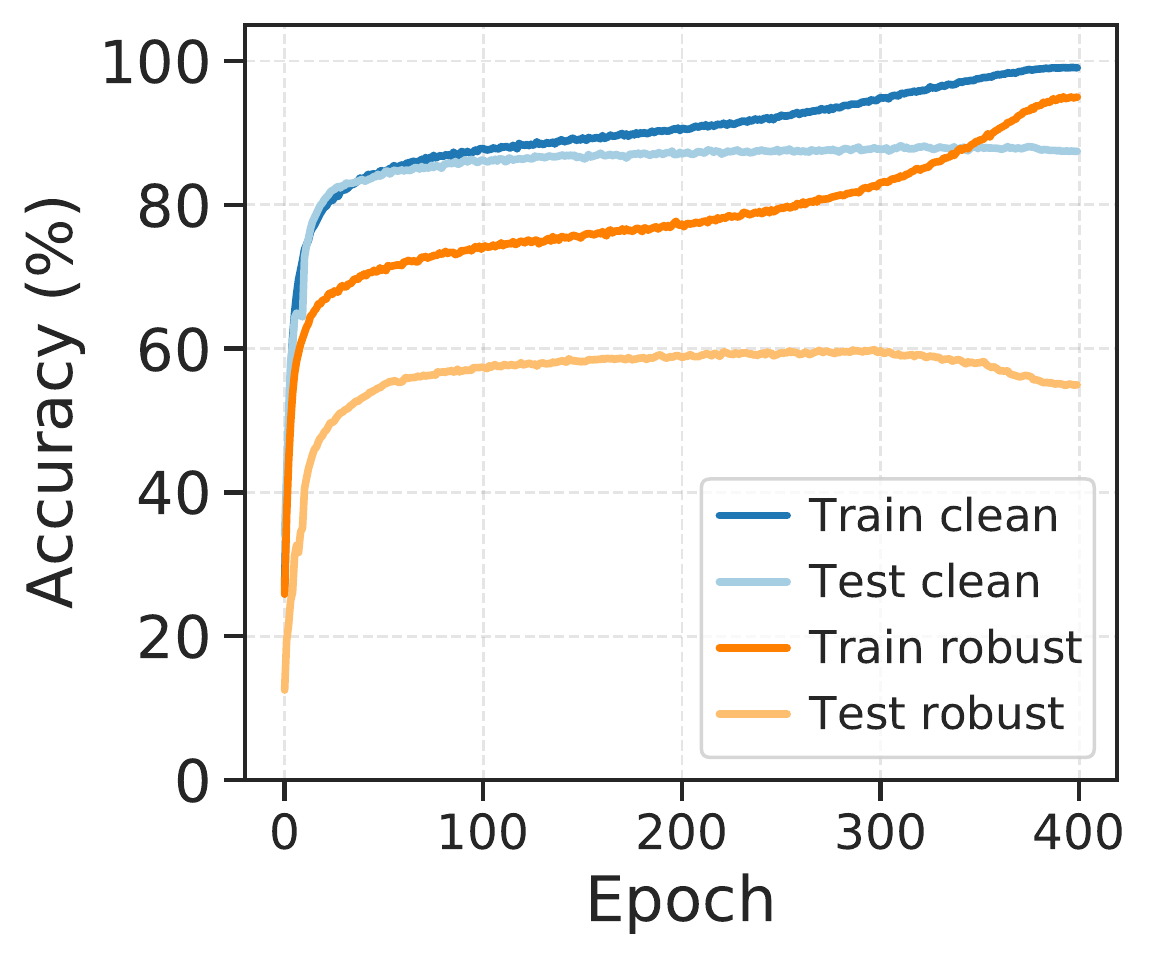}
      \vskip -0.08in
      \caption{100K generated data}
          \label{fig:amount_100k}
  \end{subfigure}
  \begin{subfigure}[t]{0.247\linewidth}
        \includegraphics[width=\linewidth]{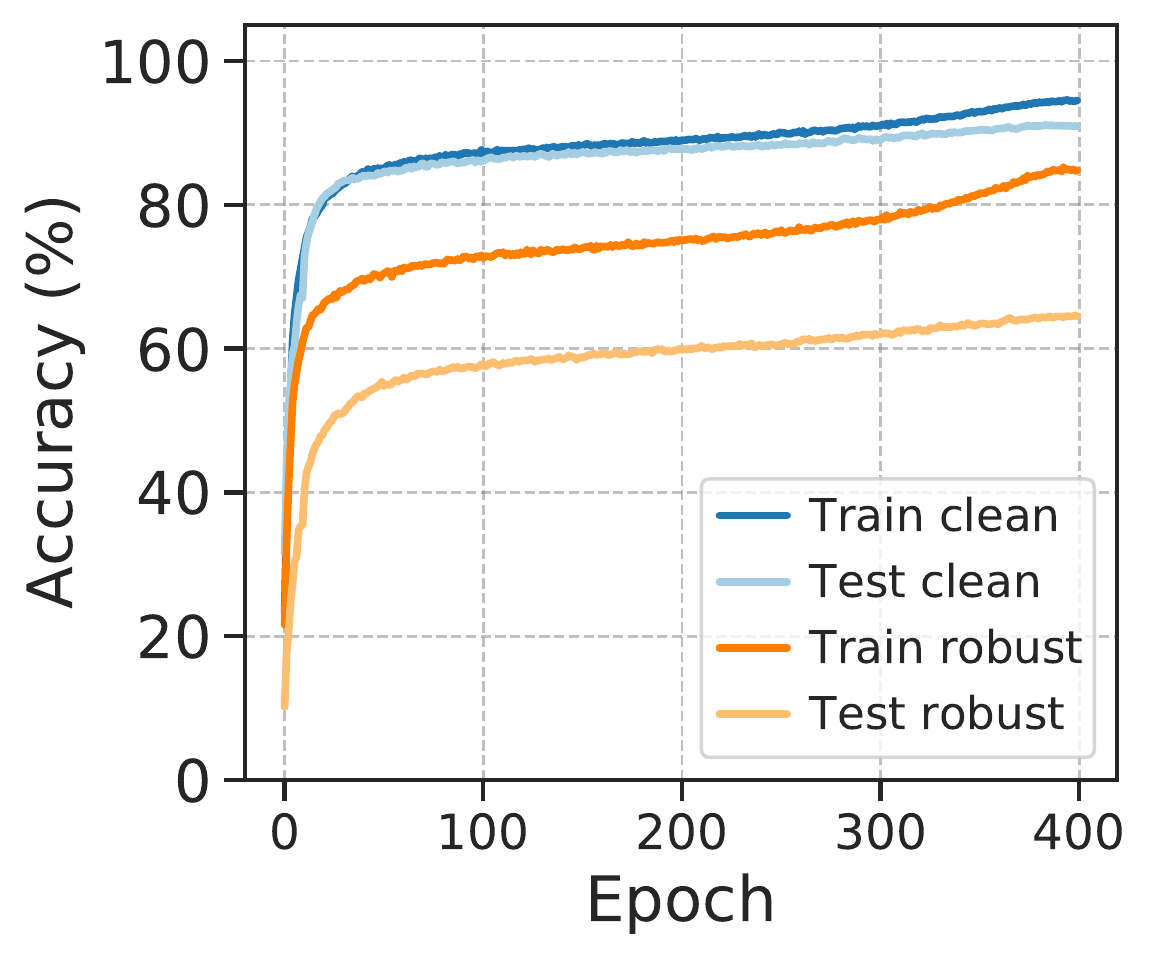}
      \vskip -0.08in
      \caption{1M generated data}
          \label{fig:amount_1m}
  \end{subfigure}
    \begin{subfigure}[t]{0.240\linewidth}
    \includegraphics[width=\linewidth]{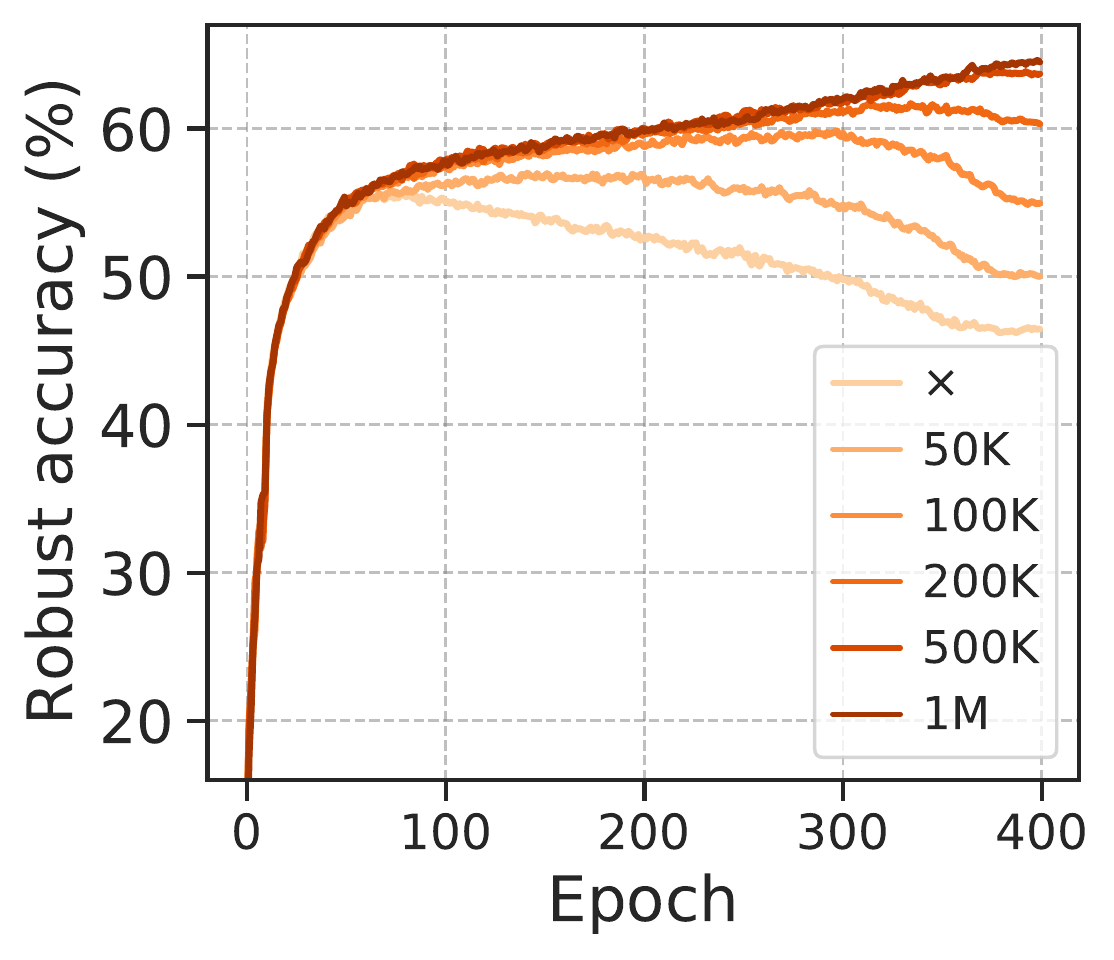}
    \vskip -0.08in
    \caption{effect of data amount}
    \label{fig:abl_amount}
  \end{subfigure}
  \vskip -0.05in
  \caption{Clean and PGD robust accuracy of AT using (a) no generated data; (b) 100K generated data; (c) 1M generated data. (d) plots the PGD test robust accuracy of AT using different amounts of generated data.}
    \label{fig:amount}
      \vspace{-0.3cm}
\end{figure*}

\subsection{Amount of Generated Data}

We can sample many more images with the generated model than we could with the original training set. According to \citet{conf/nips/GowalRWSCM21}, more DDPM generated images result in a smaller robust generalization gap. \citet{journals/corr/Rebuffi21fixing} successfully prevent robust overfitting using DDPM to generate data of fixed size (1M), achieving stable performance after a drop in the learning rate. Here we look at how the size of the generated data affects robust overfitting. The results are displayed in \cref{fig:amount} and \cref{tab:amount}.

\cref{fig:amount} (a,b,c) depicts the clean and robust accuracy on the training and test sets in relation to different amounts of EDM generated data. More results with varying data sizes are shown in \cref{app:amount}. The findings indicate that:

\begin{itemize}
    \item We can see a severe robust overfitting phenomenon when no generated data is used (\cref{fig:amount} (a), `\ding{55}' in \cref{tab:amount}). After a certain epoch, the test clean accuracy degrades slightly, whereas the `Diff' of robust accuracy is large. At the last epoch, the generalization gap between train and test robust accuracy is nearly $60\%$.

    \item Generated data can help to close the generalization gap for both clean and robust accuracy. In the final few epochs, without generated data, the training loss approaches zero. Train accuracy decreases as the size of the generated data increases, while test accuracy improves. The results show that the images generated by EDM contain those that are difficult to classify robustly, which benefits the robustness of model.
    
    \item The robust overfitting is alleviated with the increasing size of generated data, as shown in \cref{fig:amount} (d). After 500K generated images, the added generated images provide no significant improvement. In \cref{tab:amount}, `Diff' after 500K generated images become minor and `Best epoch' appears in the last few epochs. This is to be expected because the model's capacity is insufficient to utilize all of the generated data. As a result, we provide SOTA results using a large model (WRN-70-16). A longer training epoch can also aid the model's convergence on sufficient data, as seen in \cref{tab:epoch}. 
    \vspace{-0.cm}
\end{itemize}

\subsection{Data Augmentation}

Data augmentation has been shown to improve standard training generalization by increasing the quantity and diversity of training data. It is somewhat surprising that almost all previous attempts to prevent robust overfitting solely through data augmentation have failed~\citep{conf/icml/RiceWK20overfit, conf/nips/WuX020, conf/aaai/TackYJKHS22cons}. \citet{journals/corr/Rebuffi21fixing} observe that combining data augmentation with weight averaging can promote robust accuracy, but it is less effective when using external data (e.g., 80M Tiny Images dataset). While \citet{conf/nips/GowalRWSCM21} report that CutMix is compatible with using generated data, preliminary experiments in \citet{conf/icml/PangLYZY22score} suggest that the effectiveness of CutMix may be dependent on the specific implementation.

To this end, we consider a variety of data augmentations and check their efficacy for AT with generated data. Common augmentation \cite{conf/cvpr/HeZRS16padcrop} is used in image classification tasks, including padding the image at each edge, cropping back to the original size, and horizontal flipping. Cutout \cite{journals/corr/Devries17cutout} randomly drops a region of the input image. CutMix \cite{conf/iccv/YunHCOYC19cutmix} randomly replaces parts of an image with another. AutoAugment \cite{conf/cvpr/CubukZMVL19autoaug} and RandAugment \cite{conf/nips/CubukZS020randaug} employ a combination of multiple image transformations such as Color, Rotation and Cutout to find the optimal composition. IDBH \cite{conf/iclr/li23idbh}, the most recent augmentation scheme designed specifically for AT, achieves the best robust performance in the setting without additional data when compared to the augmentations mentioned above.

We consider common augmentation to be the baseline because it is the AT studies' default setting. Using 1M EDM generated data, \cref{tab:augmentation} demonstrates the performance of various data augmentations. No augmentation outperforms common augmentation in terms of robust accuracy (PGD-40 and AutoAttack). Cutout and IDBH outperform the other methods by a small margin. It should be noted that IDBH is intended for AT but also fails in the setting with generated data. In terms of clean accuracy, Cutout and AutoAugment slightly outperform common augmentation. 

To summarize, rule-based (Cutout and CutMix) and policy-based (AutoAugment,  RandAugment and IDBH) data augmentations appear to be less effective in improving robustness, particularly when using generated data. Our empirical findings contradict previous research, indicating that the efficacy of data augmentation for AT may be dependent on implementation. Thus, we use common augmentation as the default setting following \citet{conf/icml/PangLYZY22score}.

\begin{table}[t]
\centering
\small
\caption{Test accuracy (\%) with different \textbf{augmentation methods} under the ($\ell_{\infty}$, $\epsilon=8/255$) threat model on CIFAR-10, using WRN-28-10 and 1M EDM generated data.}
\vspace{0.1cm}
\renewcommand*{\arraystretch}{1.2}
    \begin{tabular}{l|ccc}
        \toprule
        Methed & Clean & PGD-40   & AA    \\ \midrule
        Common   & 91.12 & \textbf{64.61} & \textbf{63.35} \\
        Cutout    & \textbf{91.25} & 64.54 & 63.30 \\
        CutMix    & 91.08 & 64.34 & 62.81 \\
        AutoAugment   & 91.23 & 64.07 & 62.86 \\
        RandAugment   & 91.14 & 64.39 & 63.12 \\
        IDBH    & 91.08 & 64.41 & 63.24 \\
        \bottomrule
    \end{tabular}
\vspace{-0.5cm}
\label{tab:augmentation}
\end{table}%

\begin{table}[t]
\centering
\small
\caption{Test accuracy (\%) and FID with different \textbf{sampling steps} of diffusion model, under the ($\ell_{\infty}$, $\epsilon=8/255$) threat model on CIFAR-10, using WRN-28-10 and 1M EDM generated data. Here $\downarrow$ means `the lower the better'.}
\vspace{0.1cm}
\renewcommand*{\arraystretch}{1.1}
    \begin{tabular}{cccccc}
    \toprule
                                & Step & FID $\downarrow$   & Clean & PGD-40 & AA \\ \midrule
    \multirow{8}{*}{\textbf{Class-cond.}} & 5    & 35.54 & 88.92 & 57.33  & 57.78      \\
                                & 10   & 2.477 & 90.96 & 66.21  & 62.81      \\
                                & 15   & 1.848 & 91.05 & 64.56  & 63.24      \\
                                & 20   & \textbf{1.824} & \textbf{91.12}  & \textbf{64.61}  & \textbf{63.35}      \\
                                & 25   & 1.843 & 91.07 & 64.59  & 63.31      \\
                                & 30   & 1.861 & 91.10 & 64.51  & 63.25      \\
                                & 35   & 1.874 & 91.01 & 64.55  & 63.13      \\
                                & 40   & 1.883 & 91.03 & 64.44  & 63.03      \\ \midrule
    \multirow{8}{*}{\textbf{Uncond.}}     & 5    & 37.78 & 88.00 & 56.92  & 57.19      \\
                                & 10   & 2.637 & 89.40 & 62.88  & 61.92      \\
                                & 15   & 1.998 & 89.36 & 63.47  & 62.31      \\
                                & 20   & \textbf{1.963} & \textbf{89.76} & \textbf{63.66}  & \textbf{62.45}      \\
                                & 25   & 1.977 & 89.61 & 63.63  & 62.40      \\
                                & 30   & 1.992 & 89.52  & 63.51  & 62.33     \\
                                & 35   & 2.003 & 89.39  & 63.56  & 62.37     \\
                                & 40   & 2.011 & 89.44 & 63.30  & 62.24      \\ \bottomrule
    \end{tabular}
\vspace{-0.4cm}
\label{tab:step}
\end{table}%

\subsection{Quality of Generated Data}\label{fid_cifar10}

The number of sampling steps is a critical hyperparameter in diffusion models that controls generation quality and speed. Thus, we generate data with varying sampling steps in order to investigate how the quality of generated data affects model performance. We assess the quality by calculating the FID scores~\citep{conf/nips/HeuselRUNH17} computed between 50K generated images and the original training images. In \cref{app:control_fid}, we investigate the effects of various samplers and EDM formulations on model robustness.


In supervised learning, there are unconditional and class-conditional paradigms for generative modeling. Extensive empirical evidence \citep{conf/iclr/BrockDS19, conf/nips/DhariwalN21beatgan} demonstrates that class-conditional generative models are easier to train and have a lower FID than unconditional ones by leveraging data labels. We generate data in both class-conditional and unconditional settings, but give pseudo-labels in slightly different ways, to investigate the effect of class-conditioning on AT. For more information, please see \cref{app:setup}. We use EDM to generate 1M images, and the results are summarized in \cref{tab:step}.

We find that low FID of generated data leads to high clean and robust accuracy. The results in \cref{app:control_fid} come to the same conclusion. FID is a popular metric for comparing the distribution of generated images to the true distribution of data. Low FID indicates a small difference between the generated and true data distributions. The results show that we could increase the model's robustness by bringing generated data closer to the true data.

\begin{figure*}[t]
\vspace{-0.cm}
\captionof{table}{Test accuracy (\%) with different values of \textbf{batch size} (left),  \textbf{label smoothing (LS)} (middle), and \textbf{\bm{$\beta$} in TRADES} (right), under the ($\ell_{\infty}$, $\epsilon=8/255$) threat model on CIFAR-10.}
\vspace{0.1cm}
\begin{minipage}[t]{.33\linewidth}
    \begin{center}
    \renewcommand*{\arraystretch}{1.2}
        \begin{tabular}{c|ccc}
            \toprule
            Batch & \multirow{2}{*}{Clean} & \multirow{2}{*}{PGD-40} & \multirow{2}{*}{AA} \\
            Size  &                        &                         &     \\ \midrule
            128       & 91.12 & 64.77 & 63.90  \\
            256       & 91.15 & 65.76 & 64.72 \\
            512       & 91.81 & 66.15 & 65.21 \\
            1024      & 91.90  & 66.21  & 65.29 \\
            2048      & \textbf{91.98} & \textbf{66.54} & \textbf{65.50} \\
            \bottomrule
        \end{tabular}
    \end{center}
\end{minipage}
\begin{minipage}[t]{.33\linewidth}
    \begin{center}
    \renewcommand*{\arraystretch}{1.4}
        \begin{tabular}{c|ccc}
            \toprule
            LS & Clean & PGD-40   & AA    \\ \midrule
            0   & 90.40  & 64.32 & 62.83 \\
            0.1 & 91.12 & \textbf{64.61} & \textbf{63.35} \\
            0.2 & \textbf{91.23} & 64.38 & 63.27 \\
            0.3 & 91.06 & 64.35 & 63.12 \\
            0.4 & 90.82 & 64.15 & 62.87 \\
            \bottomrule
        \end{tabular}
    \end{center}
\end{minipage}
\begin{minipage}[t]{.33\linewidth}
    \begin{center}
    \renewcommand*{\arraystretch}{1.05}
        \begin{tabular}{c|ccc}
            \toprule
            $\beta$ & Clean & PGD-40   & AA    \\ \midrule
            2 & \textbf{92.46} & 63.66 & 62.32 \\
            3 & 91.83 & 64.18 & 63.03 \\
            4 & 91.30 & 64.27 & 63.11 \\
            5 & 91.12 & \textbf{64.61} & \textbf{63.35} \\
            6 & 90.77 & 64.42 & 63.23 \\
            7 & 90.39 & 64.51 & 63.29 \\
            8 & 90.25 & 64.34 & 63.19 \\
            \bottomrule
        \end{tabular}
    \end{center}
\end{minipage}
\vspace{-0.2cm}
    \label{tab:hyper}
\end{figure*}

Class-conditional generation consistently outperforms unconditional generation, with lower FID and better robust performance. With $20$ sampling steps, both settings achieve the lowest FID and best performance. Thus, for the experiments on CIFAR-10, we use class-conditional generation with $20$ sampling steps and the checkpoint provided by EDM.\footnote{https://github.com/NVlabs/edm} The additional data for baselines in \cref{sec:main_results} is generated by DDPM, which has FID of $3.28$. On CIFAR-100/SVHN datasets, we train our own model and select the model with the best FID  after $25$ sampling steps ($2.09$ for CIFAR-100, $1.39$ for SVHN, see \cref{app:fid_other_datasets}). In contrast, DDPM has FID of $5.58$ and $4.89$ on CIFAR-100 and SVHN, respectively \cite{conf/nips/GowalRWSCM21}. The large promotion on FID provides a significant performance boost over the baselines in \cref{sec:main_results}. \looseness=-1

\section{Sensitivity Analysis}\label{sec:sensitivity}

In this section, we test the sensitivity of basic training hyperparameters on CIFAR-10. WRN-28-10 models are trained for $400$ epochs using 1M data generated by EDM. $512$ is the default batch size unless otherwise specified.

\textbf{Batch size.} In the standard setting, batch size is a crucial parameter that affects the performance of the model on large-scale datasets~\citep{journals/corr/GoyalDGNWKTJH17}. In the adversarial setting without external or generated data, the batch size is typically set to $128$ or $256$. \citet{conf/iclr/PangYDSZ21bag} investigate a wide range of batch sizes, from $64$ to $512$, and find that $128$ is optimal for CIFAR-10. To evaluate the effect of batch size on sufficient data, we train the model with 5M generated images and compare its performance across five different batch sizes in \cref{tab:hyper} (left). As observed, the largest batch size of $2048$ yields the best results. It implies that robust performance is enhanced by a large batch size when sufficient training data and a fixed initial learning rate are utilized. The optimal batch size may exist when the linear scaling rule is applied~\citep{conf/iclr/PangYDSZ21bag}. A larger batch size requires additional GPU memory, but traverses the dataset more quickly. To achieve the best results in \cref{sec:main_results}, we increase the batch size based on the model size and the number of GPUs in use. We choose $2048$ batch size for WRN-28-10 on $4\times$ A100 GPUs, and $1024$ batch size for WRN-70-16 on $8\times$ A100 GPUs. 

\textbf{Label smoothing.} For standard training, label smoothing (LS) \citep{conf/cvpr/SzegedyVISW16} improves standard generalization and alleviates the overconfidence problem \citep{conf/cvpr/Hein19}, but it cannot prevent adaptive attacks from evading the model~\citep{conf/nips/TramerCBM20}. In the adversarial setting, LS is used to prevent robust overfitting without external or generated data~\citep{conf/icml/Stutz0S20, conf/iclr/ChenZ0CW21}. With 1M generated data, we evaluate the effect of LS on AT further. According to the results shown in \cref{tab:hyper} (middle), LS of $0.1$ improves both clean and robust accuracy (Clean $+0.72\%$, AA $+0.52\%$). LS of $0.2$ can further improve the clean accuracy by a small margin, but at the expense of robustness. Consistent with the findings of previous research~\citep{journals/corr/Jiang20, conf/iclr/PangYDSZ21bag}, excessive LS ($0.3$ and $0.4$) degrades the performance of the model. This is also the result of over-smoothing labels, which results in the loss of information in the output logits~\citep{conf/nips/MullerKH19}. We set $\textrm{LS}=0.1$ throughout the experiments for the best robust performance. 

\textbf{Effect of \bm{$\beta$}.} In the framework of TRADES \cite{conf/icml/ZhangYJXGJ19trades}, $\beta$ is an important hyperparameter that control the trade-off between robustness and accuracy (as in Eq.~(\ref{eq:standard_trades}) of \cref{app:setup}). As the regularization parameter $\beta$ increases, the clean accuracy decreases while the robust accuracy increases, as observed by \citet{conf/icml/ZhangYJXGJ19trades}. In contrast, when using 1M EDM generated data, the robustness of the model degrades with a large value of $\beta$. The best robustness is achieved with $\beta=5$ and smaller values of $\beta$ still contribute to improved clean accuracy. To provide the highest robust accuracy, $\beta=5$ is used on CIFAR-10/CIFAR-100. $\beta$ is set to $6$ and $8$ for SVHN and TinyImageNet, respectively.

\section{Discussion}
Diffusion models have proved their effectiveness in both adversarial training and adversarial purification; however, it is crucial to investigate how to \emph{more efficiently} exploit diffusion models in the adversarial literature. For the time being, adversarial training requires millions of generated data even on small datasets such as CIFAR-10, which is inefficient in the training phase; adversarial purification requires tens of times forward processes through diffusion models, which is inefficient in the inference phase. Our work pushes the limits on the best performance of adversarially trained models, but there is still much to explore about the learning efficiency in the future research.



\section*{Acknowledgements}
This work is supported by the National Natural Science Foundation of China under Grant 61976161, the Fundamental Research Funds for the Central Universities under Grant 2042022rc0016.



\bibliography{ref}
\bibliographystyle{icml2023}

\clearpage
\appendix
\onecolumn

\section{Technical Details}\label{app:setup}

\textbf{Adversarial training.} Let $\|\cdot\|_p$ denote $\ell_p$ norm, e.g., $\|\cdot\|_2$ and $\|\cdot\|_{\infty}$ denote the Euclidean norm $\ell_2$ and infinity norm $\ell_\infty$, respectively. $ \mathcal{B}_{p}(x,\epsilon) \coloneqq \{ x' ~|~ \| x' - x \|_{p} \leq \epsilon \}$ denotes that the input $x'$ is constrained into the $\ell_p$ ball, where $\epsilon$ is the maximum perturbation constrain. \citet{conf/iclr/MadryMSTV18pgd} formulate AT as a min-max optimization problem: 
\begin{equation}
    \label{eq:standard_at}
    \argmin_{\theta} \mathop{\mathbb{E}}\limits_{\substack{(x,y) \sim \mathcal{D}}}\left[ \max_{x'\in \mathcal{B}_{p}(x,\epsilon)} \mathcal{L}(f_{\theta}(x'), y) \right]\text{,}
\end{equation}
where $\mathcal{D}$ is a data distribution over pairs of example $x$ and corresponding label $y$, $f_{\theta}(\cdot)$ is a neural network classifier with weights $\theta$, and $\mathcal{L}$ is the loss function. The inner optimization finds adversarial example $x'$ that maximize the loss, while the outer optimization minimizes the loss on adversarial examples to update the network parameters $\theta$. 

A typical variant of standard AT is TRADES \cite{conf/icml/ZhangYJXGJ19trades}, which is applied as our AT framework. The authors show that there exists a trade-off between clean and robust accuracy and decompose Eq.~(\ref{eq:standard_at}) into clean and robust objectives. TRADES combines these two objectives with a balancing hyperparameter to control such a trade-off: 
\begin{equation}
    \label{eq:standard_trades}
    \argmin_{\theta} \mathop{\mathbb{E}}\limits_{\substack{(x,y) \sim \mathcal{D}}}\left[ \mathrm{CE}(f_{\theta}(x),y) + \beta \cdot \max_{x'\in \mathcal{B}_{p}(x,\epsilon)} \mathrm{KL}(f_{\theta}(x)\|f_{\theta}(x')) \right]\text{,}
\end{equation}
where $\mathrm{CE}$ denotes the standard cross-entropy loss, $\mathrm{KL}(\cdot\|\cdot)$ denotes the Kullback–Leibler divergence, and $\beta$ is the hyperparameter to control the trade-off. In \cref{sec:sensitivity}, we investigate the sensitivity of $\beta$ when generated data is used.

\textbf{PGD attack.} Projected gradient descent (PGD) \cite{conf/iclr/MadryMSTV18pgd} is a commonly used technique to solve the inner maximization problem in Eq. (\ref{eq:standard_at}) and (\ref{eq:standard_trades}). Let $\sgn(\bm{a})$ denote the sign of $\bm{a}$. $x_0$ is a randomly perturbed sample in the neighborhood $\mathcal{B}_{p}(x,\epsilon)$ of the clean input $x$, then PGD iteratively crafts the adversarial example for multiple gradient ascent steps $K$, formalized as:
\begin{equation}
    x_k=\clip_{x,\epsilon}(x_{k-1}+\alpha\cdot\sgn(\nabla_{x_{k-1}}\mathcal{L}(x_{k-1}, y)))\text{,}
    \label{PGD}
\end{equation}
where $x_k$ denotes the adversarial example at step $k$, $\clip_{x,\epsilon}(x')$ is the clipping function to project $x'$ back into $\mathcal{B}_{p}(x,\epsilon)$, and $\alpha$ is the step size. We will refer to this inner optimization procedure with $K$ steps as PGD-$K$. 

For adversary during AT, we apply PGD-10 attack with the following hyperparameters: for $\ell_{\infty}$ treat model, perturbation size $\epsilon=8/255$, step size $\alpha=2/255$ for CIFAR-10/CIFAR-100/TinyImageNet, and $\alpha=1.25/255$ for SVHN; for $\ell_{2}$ treat model, perturbation size $\epsilon=128/255$, step size $\alpha=32/255$ for CIFAR-10. 


\textbf{Generated data.} Here we show more details about 1M images generation on CIFAR-10/CIFAR-100. 
For the unconditional generation, we use a pre-trained WRN-28-10 to give pseudo-labels, following \citet{conf/nips/CarmonRSDL19unlabel, journals/corr/Rebuffi21fixing}. The model is standardly trained on CIFAR-10 training set and achieves 96.15\% test clean accuracy. For CIFAR-100, WRN-28-10 model achieves 80.47\% test clean accuracy. Then we sample 5M images from unconditional EDM and score each image using the highest confidence provided by the pre-trained WRN-28-10 model. We regard the class which has the highest confidence as the pseudo-label. For each class, we select the top 100K scoring images for CIFAR-10 experiments (top 10K for CIFAR-100). 

Slightly different from unconditional generating, class-conditional EDM can generate 1M samples belonging to a class directly; thus the pseudo-labels are directly determined by the class conditioning. For each class, we generate 500K and 50K images for CIFAR-10 and CIFAR-100 experiments, respectively. Similarly, we use the pre-trained WRN-28-10 model to score each image, and select the top 20\% scoring images for each class.

When generating data for the SVHN and TinyImageNet datasets, or when the amount of generated data for CIFAR-10/CIFAR-100 exceeds 1M, we directly provide pseudo-labels using class-conditional generalization. Each class has an equal number of images to maintain data balance. The number of sampling steps is set to $20$ for CIFAR-10 (\cref{fid_cifar10}),  $25$ for CIFAR-100/SVHN (\cref{app:fid_other_datasets}), and $40$ for TinyImageNet (following \citet{conf/nips/Karras22edm}).

\textbf{Datasets.} CIFAR-10 and CIFAR-100 \cite{Krizhevsky2012} consist of 50K training images and 10K test images with 10 and 100 classes, respectively. All CIFAR images are 32$\times$32$\times$3 resolution (width, height, RGB channel). SVHN \cite{Netzer2011SVHN} contains 73,257 training and 26,032 test images (0 $\sim$ 9 small cropped digits, 10 classes). TinyImageNet\footnote{http://cs231n.stanford.edu/tiny-imagenet-200.zip} contains 100K images for training, and 10K images for testing. The images are 64$\times$64$\times$3 resolution with 200 classes. 

\textbf{More about setup.} We utilize WideResNet (WRN) \cite{conf/bmvc/ZagoruykoK16wrn} following prior works \cite{conf/iclr/MadryMSTV18pgd, conf/icml/RiceWK20overfit, journals/corr/Rebuffi21fixing} which use variants of WRN family. Most experiments are conducted on WRN-28-10 (depth of 28, multiplier of 10) with 36M parameters. The experiments on WRN-28-10 are parallelly processed with four NVIDIA A100 SXM4 40GB GPUs. To evaluate how the abundant generated data affects large networks, we further use WRN-70-16 in \cref{sec:main_results}, which contains 267M parameters. We use eight A100 GPUs to train WRN-70-16. 

Following \citet{conf/icml/RiceWK20overfit} and discussion in \cref{sec:gen_data}, we perform early stopping as a defult trick. We separate first 1024 images of training set as a fixed validation set. During every epoch of AT, we pick the best checkpoint by evaluating robust accuracy under PGD-40 attack on the validation set.

\section{Additional Experiments}\label{app:add_exp}

\subsection{Original-to-Generated Ratio}\label{app:split}

\begin{figure}[h!]
\begin{center}
\vspace{0.3cm}
\begin{minipage}[h]{0.47\textwidth} 
    \centering
    \centerline{\includegraphics[width=0.88\columnwidth]{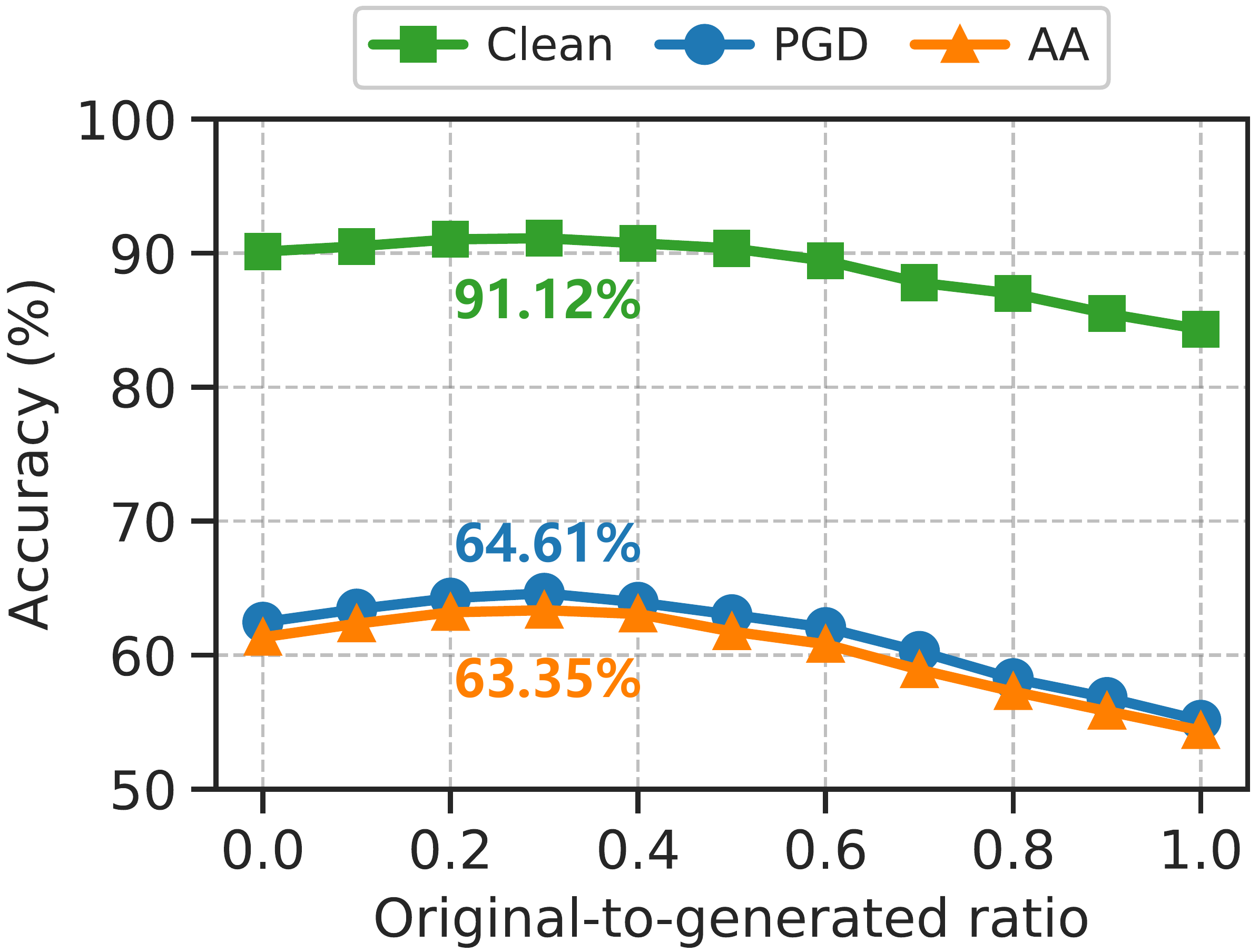}}
    \vspace{-0.2cm}
    \captionof{figure}{Clean accuracy and robust accuracy against PGD-40 and AA with respect to original-to-generated ratios (0 means generated images only, 1 means CIFAR-10 training set only). We train WRN-28-10 models against ($\ell_{\infty}$, $\epsilon=8/255$) on CIFAR-10 using 1M generated data. \label{fig:split}}
\end{minipage}
\qquad 
\begin{minipage}[h]{0.48\linewidth} 
    \centering
    \captionof{table}{Test accuracy (\%) against ($\ell_{\infty}$, $\epsilon=8/255$) on CIFAR-10 with different \textbf{original-to-generated ratio}. WRN-28-10 models are trained on 5M EDM generated data. The results of 0.2 are better than that of 0.3 consistently. \label{tab:app_split}}
    \vspace{0.1cm}
    \renewcommand*{\arraystretch}{1.1}
    \begin{tabular}{cccccc}
        \toprule
        Batch                 & Epoch                 & Ratio & Clean & PGD-40   & AA    \\ \midrule
        \multirow{2}{*}{512}  & \multirow{2}{*}{400}  & 0.2         & 91.15 & 64.97 & 64.25 \\
                              &                       & 0.3         & 91.07 & 64.88 & 64.05 \\\midrule
        \multirow{2}{*}{1024} & \multirow{2}{*}{800}  & 0.2         & 91.87 & 66.43 & 65.53  \\
                              &                       & 0.3         & 91.72 & 66.43 & 65.40 \\\midrule
        \multirow{2}{*}{2048} & \multirow{2}{*}{1600} & 0.2         & 92.16 & 67.47 & 66.34 \\
                              &                       & 0.3         & 91.88 & 67.19 & 66.29 \\
        \bottomrule
    \end{tabular}
\end{minipage}
\end{center}
\end{figure}

Original-to-generated ratio is the mixing ratio between original and generated images in the training batch, e.g., a ratio of 0.3 indicates that for every 3 original images, we include 7 generated images. We investigate how the ratio affects the performance using 1M EDM generated data. We summarize the results in \cref{fig:split}. Both clean and robust accuracy achieve the best result with the ratio of 0.3. When the generated data is greater than 1M, we pick the ratio of 0.2 to achieve better performance (\cref{tab:app_split}). We also observe that the performance of using 1M EDM generated data is better than using 50k original CIFAR-10 training set, consistent with \citet{journals/corr/Rebuffi21fixing}. The results show that generated images improve the robustness as long as the generated model can produce high-quality data.

\subsection{FID for CIFAR-100 and SVHN datasets}\label{app:fid_other_datasets}

We train our own EDM models to generate images for CIFAR-100 and SVHN experiments. The models are trained solely on CIFAR-100/SVHN training set. \cref{tab:fid_other_datasets} shows the FID scores between generated data and  CIFAR-100/SVHN training set with different sampling steps. The sampling step of 25 achieves the best FID. 
\begin{table}[h]
\centering
\small
\vspace{-0.cm}
\caption{Fr\'{e}chet inception distance (FID) between 50K EDM generated data and CIFAR-100/SVHN training set with different \textbf{sampling steps} of diffusion model. }
\vspace{0.1cm}
\renewcommand*{\arraystretch}{1.3}
    \begin{tabular}{c|cccccccc}
        \toprule
        Step & 5      & 10    & 15    & 20    & 25    & 30    & 35    & 40    \\ \midrule
        CIFAR-100  & 24.540 & 3.054 & 2.191 & 2.103 & \textbf{2.090} & 2.092 & 2.095 & 2.097 \\ 
        SVHN  & 125.765 & 5.458 & 1.661 & 1.405 & \textbf{1.393} & 1.410 & 1.428 & 1.445 \\ 
        \bottomrule
    \end{tabular}
\vspace{-0.2cm}
\label{tab:fid_other_datasets}
\end{table}%

\subsection{Ablation Studies on the Specifics of Diffusion Model Implementation}\label{app:control_fid}

We provide results on CIFAR-10 using various ODE samplers for diffusion models. We use the checkpoint for the unconditional diffusion model provided by \citet{conf/iclr/Song21scorebased} and conduct ablation studies following \citet{conf/nips/Karras22edm}. \citet{conf/iclr/Song21scorebased} employs Euler's method on ODE sampler (i.e., DDIM solver, Line 1 in \cref{tab:control_fid} (left)), whereas \citet{conf/nips/Karras22edm} discovered that Heun's second-order method (i.e., EDM solver, Lines 2 and 3) yields superior results.

For the sake of a fair comparison, we employ double sampling steps for DDIM solver because it requires only half the time of EDM solver. \cref{tab:control_fid} (left) details the time required to generate 5M images, and 1M images are selected for training. \cref{tab:control_fid} (left) demonstrates that EDM solver improves FID significantly with the same generation time as DDIM solver; additionally, EDM solver promotes both clean and robust accuracy for adversarial training. The improved selection of EDM hyperparameters can further improve the robust performance of models.

We also provide results using variance preserving (VP) and variance exploding (VE) diffusion models, originally inspired by DDPM \cite{ho2020denoising} and SMLD \cite{song2019generative}. In the implementation of EDM, the VP and VE models differ in their architecture, with DDPM++ and NCSN++ being utilized, respectively. We use VP formulation throughout the experiments in the main paper. \cref{tab:control_fid} (right) demonstrates that VE achieves a comparable FID to VP, consistent with previous findings \cite{conf/nips/Karras22edm}. Both formulations result in similarly robust model performance.

\begin{figure*}[h]
\vspace{-0.cm}
\captionof{table}{Test accuracy (\%) with different \textbf{samplers} (left) and  \textbf{EDM formulations} (right), under the ($\ell_{\infty}$, $\epsilon=8/255$) threat model on CIFAR-10. The sampling step is set to 20 in Table (right).}
\vspace{0.1cm}
\begin{minipage}[t]{.63\linewidth}
    \begin{center}
    \scalebox{0.9}{
    \renewcommand*{\arraystretch}{1.32}
        \begin{tabular}{c|cccccc}
            \toprule
            Sampler        & Step & Time (h) & FID   & Clean & PGD-40 & AA \\ \midrule
            \citet{conf/iclr/Song21scorebased} & 40   & 3.75  & 10.06 & 86.97 & 61.49  & 60.53      \\
            +Heun \& EDM $t_i$ & 20   & 3.62 & 4.03  & 87.82 & 62.17  & 61.29      \\
            +EDM $\sigma(t)$ \& $s(t)$              & 20   & 3.62 & 2.98  & 88.09 & 62.36  & 61.46     \\
            \bottomrule
        \end{tabular}}
    \end{center}
\end{minipage}
\begin{minipage}[t]{.35\linewidth}
    \begin{center}
    \renewcommand*{\arraystretch}{1.4}
        \scalebox{0.95}{
        \begin{tabular}{c|cccc}
            \toprule
             & FID   & Clean & PGD-40 & AA \\\midrule
            VP          & 1.824 & 91.12 & 64.61  & 63.35      \\
            VE          & 1.832 & 91.11 & 64.53  & 63.29     \\
            \bottomrule
        \end{tabular}}
    \end{center}
\end{minipage}
\label{tab:control_fid}
\end{figure*}

\subsection{Computational Time}\label{app:time}

We provide the runtime for class-conditional and unconditional EDM generating 5M images with different sampling steps. The generation is processed with four A100 GPUs. As shown in \cref{tab:time}, unconditional generation is faster than class-conditional one with a small margin, while resulting in lower robust performance (\cref{tab:step}). 

For adversarial training, WRN-28-10 and WRN-70-16 are parallelly processed with four and eight A100 GPUs, respectively. It takes 3.45 min on average to train one epoch on WRN-28-10 model with 2048 batch size. Training one epoch for WRN-70-16 model with 1024 batch size takes 9.93 min on average. 
\begin{table}[h]
\centering
\small
\caption{Time (h) for class-conditional and unconditional EDM generating 5M data. }
\vspace{0.1cm}
\renewcommand*{\arraystretch}{1.3}
    \begin{tabular}{c|cccccccc}
        \toprule
        Step              & 5    & 10   & 15   & 20    & 25    & 30    & 35    & 40    \\ \midrule
        Class-conditional & 2.59 & 5.37 & 8.13 & 10.92 & 13.72 & 16.47 & 19.26 & 22.03 \\ 
        Unconditional     & 2.52 & 5.30  & 8.05 & 10.81  & 13.57 & 16.40  & 19.18 & 21.95 \\
        \bottomrule
    \end{tabular}
\vspace{-0.2cm}
\label{tab:time}
\end{table}%

\clearpage
\subsection{Amount of Generated Data}\label{app:amount}

\begin{figure}[htbp]
\begin{center}
\centerline{\includegraphics[width=\columnwidth]{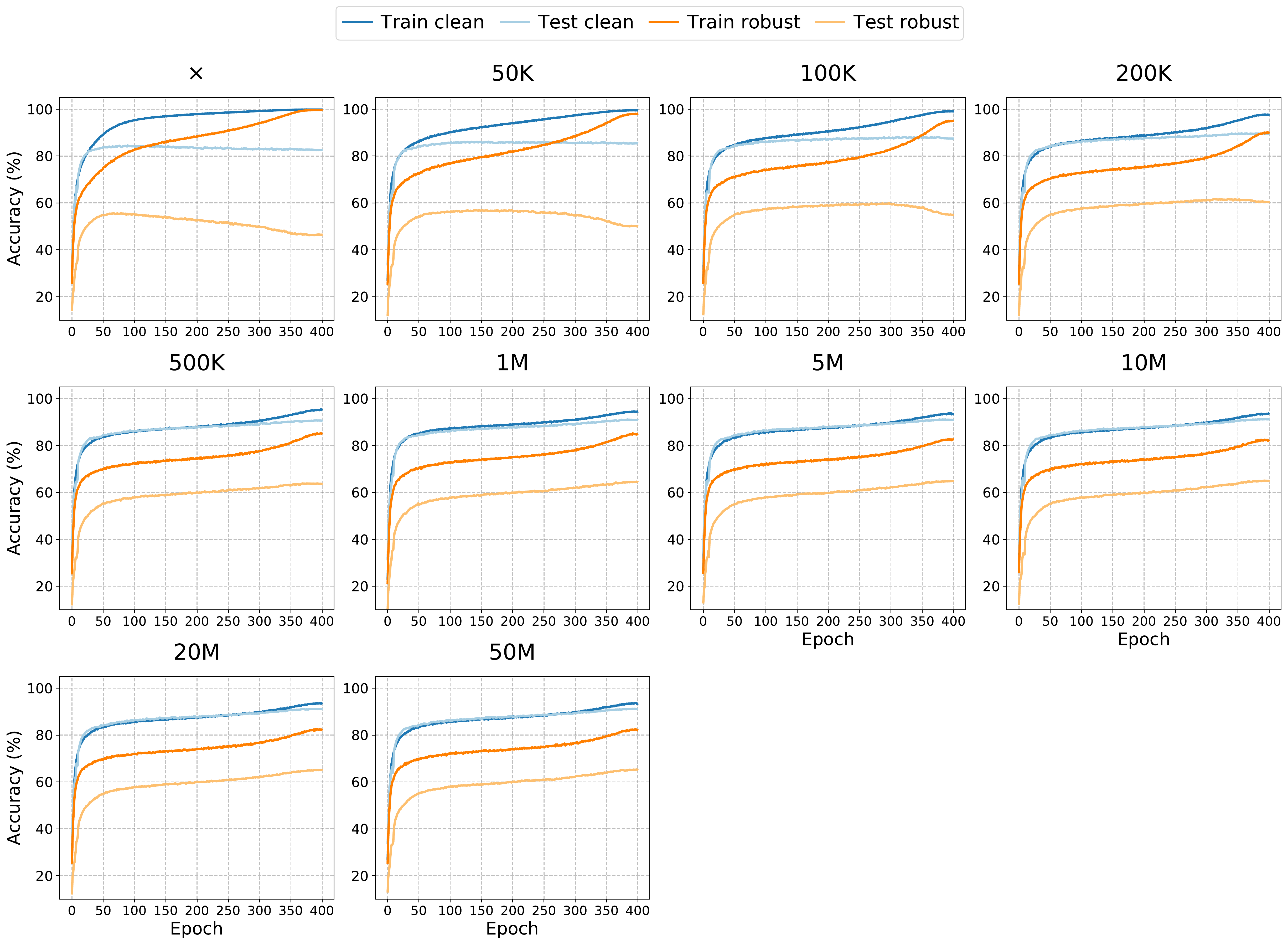}}
\vskip -0.1in
\caption{Clean and PGD robust accuracy of adversarial training using different amounts of generated data. }
\label{fig:app_amount}
\end{center}
\vskip -0.2in
\end{figure}

\cref{fig:app_amount} shows clean and PGD robust accuracy using different amounts of generated data. The robust overfitting is alleviated significantly as the size of generated data increases. After 500K generated images, the added generated images can not further close the generalization gap for both clean and robust accuracy. This is expected as the model capacity is too low to take advantage of all this generated data.

\end{document}